\documentclass[sn-mathphys,Numbered]{sn-jnl}

\usepackage{graphicx}%
\usepackage{multirow}%
\usepackage{amsmath,amssymb,amsfonts}%
\usepackage{amsthm}%
\usepackage{mathrsfs}%
\usepackage[title]{appendix}%
\usepackage{xcolor}%
\usepackage{textcomp}%
\usepackage{manyfoot}%
\usepackage{booktabs}%
\usepackage{algorithm}%
\usepackage{algorithmicx}%
\usepackage{algpseudocode}%
\usepackage{lineno,hyperref}
\usepackage{tikz}
\usepackage{tikz-qtree}
\usepackage{pgfplots}
\pgfplotsset{width=7cm,compat=1.15}
\usepgfplotslibrary{statistics}
\usepackage{listings}
\usepackage[most]{tcolorbox}
\usepackage{inconsolata}

\theoremstyle{thmstyleone}%
\theoremstyle{thmstyletwo}%

\theoremstyle{thmstylethree}%

\begin{document}

\title[Context-Aware Semantic Similarity Measurement for Unsupervised Word Sense Disambiguation]{Context-Aware Semantic Similarity Measurement for Unsupervised Word Sense Disambiguation}


\author{\fnm{Jorge} \sur{Martinez-Gil}}\email{jorge.martinez-gil@scch.at}

\affil{\orgname{Software Competence Center Hagenberg}, \orgaddress{\street{Softwarepark 32a}, \city{Hagenberg}, \postcode{4232}, \country{Austria}}}


\abstract{The issue of word sense ambiguity poses a significant challenge in natural language processing due to the scarcity of annotated data to feed machine learning models to face the challenge. Therefore, unsupervised word sense disambiguation methods have been developed to overcome that challenge without relying on annotated data. This research proposes a new context-aware approach to unsupervised word sense disambiguation, which provides a flexible mechanism for incorporating contextual information into the similarity measurement process. We experiment with a popular benchmark dataset to evaluate the proposed strategy and compare its performance with state-of-the-art unsupervised word sense disambiguation techniques. The experimental results indicate that our approach substantially enhances disambiguation accuracy and surpasses the performance of several existing techniques. Our findings underscore the significance of integrating contextual information in semantic similarity measurements to effectively manage word sense ambiguity in unsupervised scenarios. The source code of this approach is available at: \url{https://github.com/jorge-martinez-gil/uwsd}.}

\keywords{Natural Language Processing, Knowledge Engineering, Semantic Similarity Measurement}



\maketitle

\section{Introduction}
Semantic similarity refers to the extent to which two text pieces convey the same meaning \cite{key-Navigli}. Traditional semantic similarity measurement strategies rely on various approaches considering the overlap of features between the two texts to be compared \cite{key-Lastra-HESML}. However, these approaches suffer from several limitations, as they fail to consider the context in which the words and sentences are used. In other words, in conventional semantic similarity measures, the resemblance between two entities is based on their definitions, relationships, and other linguistic or extrinsic features \cite{key-martinez-mlwa}. However, in real-world applications, the context in which entities are being compared can affect their resemblance.

Furthermore, word sense ambiguity is typical in natural language processing (NLP) because words often have numerous meanings depending on their context. Word sense disambiguation (WSD) aims to identify the correct meaning of a word in a given context \cite{navigli2009word}. While supervised WSD approaches have achieved high accuracy, they are limited by the availability of annotated data. In contrast, unsupervised approaches rely on something other than annotated data but often suffer from lower accuracy due to that lack of supervision.

This research proposes a Context-Aware Semantic Similarity (CASS) measurement approach for unsupervised WSD to overcome the accuracy of the results traditionally achieved through unsupervised strategies. Our strategy incorporates contextual information into the similarity measurement process to reduce language ambiguity. This approach allows us to automatically identify the most likely sense of an ambiguous word based on its context without relying on annotated data. In this way, CASS can improve the results in many domains where relevance may vary based on the context.

Our strategy is not the first in that direction since several unsupervised WSD techniques have been proposed \cite{key-Han,key-Moradi,key-Rahman,key-Ustalov}. However, these methods use different strategies to consider the specific context in which the words appear. Our proposed strategy addresses current limitations using a novel CASS that adequately incorporates contextual information into the disambiguation process. Therefore, the primary contributions of this research can be summarized as follows:

\begin{itemize}
\item A novel disambiguation strategy that considers the context in which words are used to improve the accuracy and relevance of language models beyond the traditional methods to identify synonyms. The proposed approach for unsupervised WSD can benefit languages with limited annotated data, whereas supervised approaches may not be as effective. 
\item Evaluation of the proposed method on a complete benchmark dataset and comparison of its performance with several state-of-the-art unsupervised WSD techniques. The experimental results show that the method improves disambiguation accuracy and outperforms several existing techniques, especially when annotated data is limited or unavailable.
\end{itemize}

The remainder of this paper is organized as follows. Section 2 provides an overview of related work in unsupervised WSD. Section 3 introduces the problem statement of this research. Section 4 presents the details of the proposed CASS measurement approach. In Section 5, the experimental setup is described, and the evaluation results are presented. Section 6 discusses the results obtained and directions for future work. Finally, the paper concludes in Section 7.

\section{State-of-the-art}
This section presents an overview of CASS measurement, discussing its objectives and methods. We examine the state-of-the-art strategies used for this task and discuss their challenges and limitations, such as the difficulty of capturing context-dependent nuances and the lack of a universally accepted evaluation methodology. Finally, we highlight some of the potential applications.

\subsection{Semantic Similarity}
Semantic similarity is an essential concept in NLP that has been extensively studied in the literature \cite{key-sts,key-Harispe,key-Lastra3,key-Lastra,key-martinez-kbs,key-sematch}. Traditional approaches to measuring semantic similarity are typically based on the study of intrinsic characteristics of the words (lexical methods) or their distribution in sufficiently meaningful text corpora (distributional semantics) \cite{key-martinez-jfis}. Lexical methods rely on the meaning of individual words and their relationships to each other \cite{key-martinez-eswa2}. In contrast, distributional semantics techniques aim to capture the meaning of words based on their co-occurrence patterns in large corpora \cite{key-Bollegala}.

CASS is a family of NLP techniques that measures the semantic similarity between two words or phrases in a given context. This family has gained increasing attention in recent years due to its ability to capture not just the meaning of words but additional nuances by considering the context in which they are used \cite{key-Pilehvar}. It represents an extension of traditional semantic similarity measurement since the latter does not consider the text's context. Recent advances have seen the development of CASS methods that consider the context in which words are used. 

In addition, a universally accepted evaluation methodology for assessing the performance of CASS measures is needed. This lack of consensus makes it difficult to compare the best approaches for a variety of applications, which slows down overall progress in the field. In the research that has been done, numerous evaluation measures have been suggested. However, they frequently have drawbacks, such as favoring particular data. Therefore, additional research is required in order to develop a methodology for evaluation that is both comprehensive and objective.

\subsection{Applications}
CASS measures are intended to increase text understanding capabilities. This can be especially helpful in various applications, including web search, document classification, question-answering, and text summarization. CASS measures the semantic similarity between two words, and context plays a vital role in determining this similarity. Search engines use semantic similarity to retrieve documents that match the meaning of the user's query. In document classification, understanding the document's meaning is essential, while understanding the question's meaning is crucial in question-answering systems. Text summarization involves condensing text into a shorter version and retaining essential information. CASS can improve accuracy and relevance in these applications by capturing language nuances and context.

\subsection{Word Sense Disambiguation}
CASS measurement and WSD are related concepts in the field of NLP, but they differ significantly. Semantic similarity measurement involves determining how similar two words or phrases are in terms of meaning \cite{key-chandrasekaran}. CASS measurement considers the context in which the words or phrases appear in a sentence or document and their inherent semantic properties. This approach can help capture nuances and subtleties in meaning that might be missed by other methods that rely solely on the intrinsic properties of the words or phrases.

WSD, conversely, is determining which word's meaning is intended in a particular context \cite{key-apidianaki}. This is particularly important for words with multiple meanings \cite{key-loureiro2022lmms}. WSD can be a challenging problem, especially when the context is ambiguous or there are few clues to help distinguish among the possible senses \cite{key-eyal}. The challenge of unsupervised WSD is also essential, as evidenced by several recent papers providing ideas for meeting the challenge when adequate training datasets are unavailable \cite{key-Han,key-Moradi,key-Rahman,key-Ustalov}. Therefore, CASS and WSD are essential tools in NLP, but they have different goals and use cases. 

\subsection{Contribution over the state-of-the-art}
The training-test gap for models based on unsupervised language modeling makes it difficult for these methods to compute semantic similarity and perform word sense disambiguation correctly. Existing annotated datasets are typically small, making it challenging to train supervised neural models. Our proposed strategy, which incorporates CASS, is the foundation of our contribution to the state-of-the-art since it alleviates the problem in scenarios where appropriate training datasets are unavailable. We have tested our strategy against the most recent WSD benchmark dataset and discovered that it performs better in accuracy than other methods. Furthermore, our approach is appropriate for large-scale applications and information retrieval because it is computationally effective and scalable. Our work thus advances the development of CASS methodologies and shows how context integration can increase the accuracy and robustness of WSD techniques.

\section{Problem Statement}
There are several strategies for calculating CASS. Each CASS strategy has particular strengths and limitations, and the choice depends on the specific scenario to be faced. However, it is generally possible to partition and address problem in several steps as we will see below.

\subsection{Context-Aware Semantic Similarity Measurement}
The problem that we address here can be formulated as follows: Let $\mathcal{C}$ be the set of contexts, $\mathcal{W}$ be the set of words, and $\mathcal{S}$ be the set of semantic similarity scores between word pairs.

Given a context $c \in \mathcal{C}$ and two words $w_1, w_2 \in \mathcal{W}$, the task is to compute a CASS score $\mathcal{S}(c, w_1, w_2) \in \mathcal{S}$ between the word pair $w_1, w_2$ in the context $c$.

This can be formally developed as a mathematical expression as in Eq. \ref{eq:cass}:

\begin{equation}
\mathcal{S}(c, w_1, w_2) = f(c, w_1, w_2)
\label{eq:cass}
\end{equation}

where $f$ is the function that maps a context $c$ and two words $w_1$ and $w_2$ to a semantic similarity score.

Therefore, the goal is to find a function $f$ that considers the context $c$ for an accurate calculation of the semantic similarity score $\mathcal{S}(c, w_1, w_2)$. 

\subsection{Word Sense Disambiguation}
Let $\mathcal{W}$ be a set of words with more than one sense, and let $\mathcal{S}$ be a set of senses associated with each word in $\mathcal{W}$. Let $\mathcal{C}$ be a text corpus consisting of a set of documents $\mathcal{D} = \{d_1, d_2, ..., d_n\}$. For each word $w$ in $\mathcal{W}$, let $\mathcal{T}(w)$ be the set of occurrences of $w$ in $\mathcal{C}$, and let $\mathcal{S}(w)$ be the set of senses associated with $w$.

The goal of WSD is to assign a sense $s$ in $\mathcal{S}(w)$ to each occurrence $t$ in $\mathcal{T}(w)$, such that the assigned sense is the most appropriate for the context in which $t$ appears.

Formally, let $\mathcal{S}'(w) = \{s_1, s_2, ..., s_m\}$ be a set of candidate senses for $w$. For each occurrence $t$ in $\mathcal{T}(w)$, we seek to find the sense $s$ in $\mathcal{S}'(w)$ that maximizes the probability $\mathcal{P}(s | t, C)$, where $\mathcal{C}$ is the context in which $t$ appears as expressed in Eq. \ref{eq:wsd}.

\begin{equation}
s* = argmax_s \mathcal{P}(s | t, C)
\label{eq:wsd}
\end{equation}

where $s*$ is the assigned sense for $t$, and $argmax_s$ denotes the sense that maximizes the probability.

Several approaches to estimating the probability $\mathcal{P}(s | t, C)$ include supervised learning, unsupervised learning, and knowledge-based methods. In supervised learning, an annotated dataset of word occurrences with their corresponding senses is used to train a classifier that predicts the sense for new occurrences. In unsupervised learning, clustering or probabilistic models usually group similar word occurrences into clusters, each representing a sense. In knowledge-based methods, external knowledge sources, such as dictionaries or semantic networks, infer the most appropriate sense for a given word occurrence.

\section{Methods}
In our research, we aim to tackle the challenge of word ambiguity, which refers to the problem of words having multiple meanings based on the context in which they are used. We propose adapting existing methods incorporating contextual information to address this issue.

Among the current state-of-the-art methods for contextual language processing, we identified four stand-out approaches: BERT \cite{key-Bert}, ELMo \cite{key-elmo}, USE \cite{key-cer}, and WMD \cite{key-kusner}. Each method employs a unique technique for capturing contextual information, making them suitable for different use cases \cite{key-martinez-dawak}.

To adapt these methods for addressing the issue of word ambiguity, we propose using them to create contextualized embeddings. This involves representing each text unit as a vector that considers the context in which it appears. For instance, BERT (Bidirectional Encoder Representations from Transformers) \cite{key-Bert} is a deep neural network that uses a transformer architecture to generate contextualized word embeddings. Similarly, ELMo (Embeddings from Language Models) \cite{key-elmo} creates embeddings by training bidirectional models on large text corpora.

On the other hand, USE (Universal Sentence Encoder) \cite{key-cer} is a pre-trained encoder that can be used to generate sentence embeddings that capture the contextual meaning of a sentence. Lastly, WMD (Word Mover's Distance) \cite{key-kusner} is a distance-based metric that calculates the similarity between two documents based on the distance between their constituent words.

Adapting these strategies to capture contextual details might improve the performance of tasks that involve disambiguating words. Our contribution is to show how existing strategies can be adapted for dealing with the problem of word ambiguity, which has important implications for a wide range of applications as we have already seen.

\subsection {Definition of our method for Context-Aware Semantic Similarity}
Given a word $w$, a context $C$, and an exclusion list $E$, the function $\texttt{CASS}(w, C, E)$ should find a synonym $s^*$ of $w$ that, when substituted in $C$, results in the slightest alteration of the meaning of $C$. The process is defined as follows:

Let $W = \{s_1, s_2, \ldots, s_n\}$ be the set of synonyms of $w$, excluding any synonyms that are contained in $E$. The function transforms $C$ and $C_{s_{i}}$, where $C_{s_{i}}$ is the context $C$ with $w$ replaced by $s_i$, into embeddings using a pre-trained transformer model. These embeddings are multi-dimensional vectors, $E(C) \in \mathcal{R}^d$ and $E(C_{s_i}) \in \mathcal{R}^d$, where $d$ is the dimension.

The objective is to find $s^* \in W$ that minimizes the semantic change or distance from $C$ to $C_{s_i}$, which is inverse to the cosine similarity between their embeddings:

\begin{equation}
    s^* = \arg\min_{s_i \in W} \left( 1 - \frac{E(C) \cdot E(C_{s_i})}{\|E(C)\| \|E(C_{s_i})\|} \right)
\end{equation}

where $\cdot$ denotes the dot product and $\|\cdot\|$ denotes the Euclidean norm.

Next, we will see the alternatives to build the embeddings using existing pre-trained transformer models.

\subsection {BERT embeddings}
One prevalent method that could be used for CASS is based on BERT embeddings \cite{key-Bert}. BERT embeddings are vector representations of words or sentences in a high-dimensional space learned from large text corpora. BERT embeddings are context-aware since they capture the meaning of text based on their surrounding context.

Let $x_1, x_2, ..., x_n$ be a succession of input tokens defining a sentence, and let $h_i$ be the contextualized model for the $i$-th token obtained using the BERT model.

We can obtain the sentence-level embedding $S$ by taking a weighted average of the token embeddings as in Eq. \ref{eq:bert1}:

\begin{equation}
S = \frac{1}{n} \sum_{i=1}^{n} \alpha_i h_i
\label{eq:bert1}
\end{equation}
 
where $\alpha_i$ is the weight assigned to the $i$-th token, and is given by Eq. \ref{eq:bert2}:

\begin{equation}
	\alpha_i = \frac{\exp(w^T h_i)}{\sum_{j=1}^{n} \exp(w^T h_j)} \
	\label{eq:bert2}
\end{equation}
 
Here, $w$ is a parameter vector that defines the importance of each token in the sentence. Note that the weights $\alpha_i$ are learned during training and are used to give higher importance to the most relevant tokens. 

\subsection{ELMo}
ELMo is a deep contextualized word representation model using a bi-directional language (biLM) to generate word embeddings \cite{key-elmo}. The biLM is trained on a large corpus of text data to predict the next word in a sequence of words given the previous words in both forward and backward directions.

Combining the hidden states of the biLM at each layer allows the production of the ELMo representation of a word. Let us denote the biLM as a function $f_{biLM}(x)$ that takes a sequence of words $x$ as input and produces a set of hidden states $H = {h_1, h_2, ..., h_L}$ at each layer $l$.

The ELMo representation of a sentence $s_i$ is then computed as a weighted sum of the hidden states at each layer $L$ as in Eq. \ref{eq:elmo}:

\begin{equation}
ELMo^{s} = \gamma^{s}\left[\sum{j=0}^{L-1}s_{j}\cdot\mathbf{w}{j}\right] + \gamma^{s}{x}\left[\sum_{j=0}^{L-1}\sum_{k=1}^{T_{j}}s_{j,k}\cdot\mathbf{w}_{j,k}\right]
\label{eq:elmo}
\end{equation}

where ELMo represents the embedding for a given sentence $s$, $L$ is the number of layers in the ELMo model, $T{j}$ is the number of tokens in the $j$-th layer, $s_{j}$ and $s_{j,k}$ are the activations of the $j$-th layer for the sentence and the $k$-th token in the $j$-th layer, respectively, $\mathbf{w}{j}$ and $\mathbf{w}{j,k}$ are the weights for the $j$-th layer and the $k$-th token in the $j$-th layer, and $\gamma^{s}$ and $\gamma^{s}_{x}$ are scalar weights obtained during training.

The weights capture the importance of each layer for the specific task and allow ELMo to generate context-dependent embeddings that are useful for our purposes.

\subsection {Universal Sentence Enconder}
Let us say we have two sentences $\mathcal{X}$ and $\mathcal{Y}$, and we want to calculate their CASS using the USE embeddings \cite{key-cer}. We first obtain the USE embeddings of $\mathcal{X}$ and $\mathcal{Y}$, denoted by $\mathbf{e_X}$ and $\mathbf{e_Y}$, respectively.

The semantic similarity $ss$ between $\mathbf{e_X}$ and $\mathbf{e_Y}$ is then calculated as in Eq. \ref{eq:use}:

\begin{equation}
\text{ss} = \frac{\mathbf{e_\mathcal{X}} \cdot \mathbf{e_\mathcal{Y}}}{\lVert \mathbf{e_\mathcal{X}} \rVert \cdot \lVert \mathbf{e_\mathcal{Y}} \rVert}
\label{eq:use}
\end{equation}

Where:

$\cdot$ denotes the dot product between the embeddings and
$\lVert \cdot \rVert$ denotes the L2 norm of the embeddings, i.e., the length of the embedding vectors

If necessary, Eq. \ref{eq:use2} can also work with items from the sentences.

\begin{equation}
\text{ss} = \frac{\sum_i(e_{\mathcal{X}i} \cdot e_{\mathcal{Yi})}}{\sqrt{\sum_i(e_{\mathcal{X}i}^2)} \cdot \sqrt{\sum_i(e_{\mathcal{Y}i}^2)}}
\label{eq:use2}
\end{equation}

Where:

$e_{Xi}$ and $e_{Yi}$ are the $i^{th}$ elements of the embeddings $\mathbf{e_X}$ and $\mathbf{e_Y}$, respectively.

\subsection {Word Mover's Distance}
The Word Mover's Distance (WMD) measures the semantic similarity between two texts, which considers the distances between the individual words in the texts \cite{key-kusner}. The mathematical formulation of the WMD can be described as follows:

Let $\mathcal{D}$ be a metric space of word embeddings, and let be $\mathcal{X}$ and $\mathcal{Y}$ two sentences of $n$ and $m$ words, respectively. We also have a matrix $T$, which tells us how much of a word in $\mathcal{X}$ moves to a word in $\mathcal{Y}$, and this is represented by a non-negative number in $T_{ij}$. The cost of moving from one word to another is represented by $c(i,j)$, which is the distance between the word $i$ and word $j$. We need to make sure that the total flow from each word in $\mathcal{X}$ is equivalent to the value of $\mathcal{X}_i$, which can be achieved by setting $\sum_{j}T_{ij} = \mathcal{X}_i$. With these constraints in mind, we can use Eq. \ref{eq:wmd} to find the minimum cumulative cost of transforming $\mathcal{X}$ into $\mathcal{Y}$.

\begin{equation}
	  \begin{gathered}
	arg \ min \			\sum_{i,j=1}^{n}T_{ij}c(i, j) \\
	subject \ to \ 	\sum_{j=1}^{n}T_{ij} = \mathcal{X}_i \ \forall i \in \{1, 2, 3 \cdots n\} \ \wedge 
			\	\sum_{i=1}^{n}T_{ij} = \mathcal{Y}_j \ \forall j \in \{1, 2, 3 \cdots n\}
  \end{gathered}
	\label{eq:wmd}
\end{equation}

A wide range of word embeddings can be used here, e.g., word2vec \cite{key-Mikolov}. Furthermore, the optimization problem can be solved using linear programming techniques. The resulting WMD measures the semantic similarity between $\mathcal{X}$ and $\mathcal{Y}$, considering the distances between the individual words in the documents. The WMD has outperformed traditional bag-of-words and vector space models in the past \cite{skianis2020boosting}.

\section{Results}
Here, we showcase the results of our WSD experiments. Through a detailed analysis of various embedding models, we have compared the outcomes of our proposed strategy with commonly employed strategies to measure their impact.

\subsection{Empirical Setup and Baseline Selection} 
Our research proposes a CASS measurement method for unsupervised WSD. The proposed method measures the semantic similarity between a target word and its candidate senses based on the context in which the target word appears. We compare our approach with several unsupervised techniques for each use case in the dataset. The experiments are tested on a computer with 32 GB of RAM and an i7-8700 CPU running at 3.20 GHz on Windows 10.

We will use two baselines here, one weak and one strong. The weak baseline (\textit{Random Option}, RO) calculates the probability of giving a correct answer randomly. So, in cases where two possible alternatives are considered, there would be a probability of 50\%, in case of three, 33.33\%, and so on. The strong baseline is one of the most commonly used baselines for WSD; the \textit{Most Frequent Sense }(MFS) method. The MFS baseline assigns the most frequent sense of a word in a given dataset to all instances of that word. It is a method that is difficult to replicate in the real world by a computer because it requires external knowledge (i.e., the most frequent sense for a given word). However, it is a natural solution for people.

We must compute the most frequent sense of each target word in the training data to implement this strong baseline. Then, we assign the most frequent sense as the predicted sense for each instance of the target word in the test data. While this strong baseline is very simple, the literature shows that it can be surprisingly effective, especially for words with a highly dominant sense.

\subsection{Dataset}
In this work, we are working with the CoarseWSD-20 dataset \cite{key-loureiro}, which is a dataset for figuring out the actual meaning of words that, in practice, can have different meanings. The dataset is made from Wikipedia and only includes nouns. It focuses on 20 words that can have 2 to 5 different meanings. The dataset contains 10,196 cases, which helps test WSD models, as it has all the senses in the test sets. This makes it particularly suitable for evaluating WSD models.

As our method is fully unsupervised, we do not need to use the training instances offered. At the same time, if we were to compete with solutions that use such training samples, the comparison would be unfair. So, we will limit ourselves to the comparison with other unsupervised techniques, particularly the weak (RO) and the strong (MFS) baselines. In addition, we need to adapt some mapping classes to facilitate the disambiguation slightly.

\subsection{Evaluation Criteria}
We use a standard evaluation metric called accuracy to evaluate the performance of different WSD models on the CoarseWSD-20 dataset. Accuracy is the proportion of correctly identified senses from the total number of instances in the test set. In addition, we will break down the results by use case and globally. Both for our approach and for the baselines we compare with.

\subsection{Empirical Evaluation}
We aim to evaluate the proposed CASS measurement method for unsupervised WSD and compare it with several unsupervised WSD methods. The solid blue color represents the results obtained through our strategy. The black color represents the weak baseline (the results could be replicated by selecting a random option). In contrast, the red represents the strong baseline (the results could be replicated with external knowledge about the most frequently used sense).

\begin{figure}
\begin{tikzpicture}
\begin{axis}[bar width=10pt,    ylabel={Accuracy},    xlabel={Category},    xtick=data,    xticklabels={        apple,        arm,        bank,        bass,        bow,        chair,        club,        crane,        deck,        digit    },    ymin=0,    ymax=100,    width=\textwidth,    height=8cm,    legend style={at={(0.5,-0.15)},      anchor=north,legend columns=-1},    enlarge x limits=0.03, xticklabel style={rotate=45 }]

\addplot [ybar, fill=blue!30] coordinates {
    (1,87.01)
    (2,92.68)
    (3,99.12)
    (4,72.18)
    (5,53.02)
    (6,53.07)
    (7,62.87)
    (8,89.81)
    (9,86.87)
    (10,78.57)
};

\addplot[ybar,mark=*,red] coordinates {
    (1,61.43)
    (2,73.78)
    (3,95.16)
    (4,72.20)
    (5,54.4)
    (6,67.7)
    (7,53.46)
    (8,51.59)
    (9,86.8)
    (10,78.57)
};

\addplot[ybar,mark=*,black]  coordinates {
    (1,50)
    (2,50)
    (3,50)
    (4,33.33)
    (5,33.33)
    (6,50)
    (7,33.33)
    (8,50)
    (9,50)
    (10,50)
};
\end{axis}
\end{tikzpicture}

\begin{tikzpicture}
\begin{axis}[ bar width=10pt,    ylabel={Accuracy},    xlabel={Category},    xtick=data,    xticklabels={        hood,        java,        mole,        pitcher,        pound,        seal,        spring,        square,        trunk,        yard    },    ymin=0,    ymax=100,    width=\textwidth,    height=8cm,    legend style={at={(0.5,-0.55)},      anchor=north,legend columns=-1},    enlarge x limits=0.03, , xticklabel style={rotate=45 }   ]
\addplot [ybar, fill=blue!30] coordinates {
    (1,79.27)
    (2,61.17)
    (3,66.02)
    (4,96.00)
    (5,86.60)
    (6,52.62)
    (7,60.17)
    (8,57.97)
		(9,41.56)
    (10,81.95)
};

\addplot[ybar,mark=*,red] coordinates {
    (1,57.31)
    (2,61.17)
    (3,37.37)
    (4,99.5)
    (5,89.7)
    (6,36.08)
    (7,51.64)
    (8,49.75)
    (9,61.02)
    (10,84.7)
};

\addplot[ybar,mark=*,black]  coordinates {
    (1,33.33)
    (2,50)
    (3,20)
    (4,50)
    (5,50)
    (6,25)
    (7,33.33)
    (8,25)
    (9,33.33)
    (10,50)
};
\end{axis}
\end{tikzpicture}
\caption{Results obtained for the CoarseWSD-20 dataset using UWSD+BERT. The solid blue bar represents the results obtained. While the black and red colors represent the weak and strong baselines, respectively}
\label{fig:bert}
\end{figure}
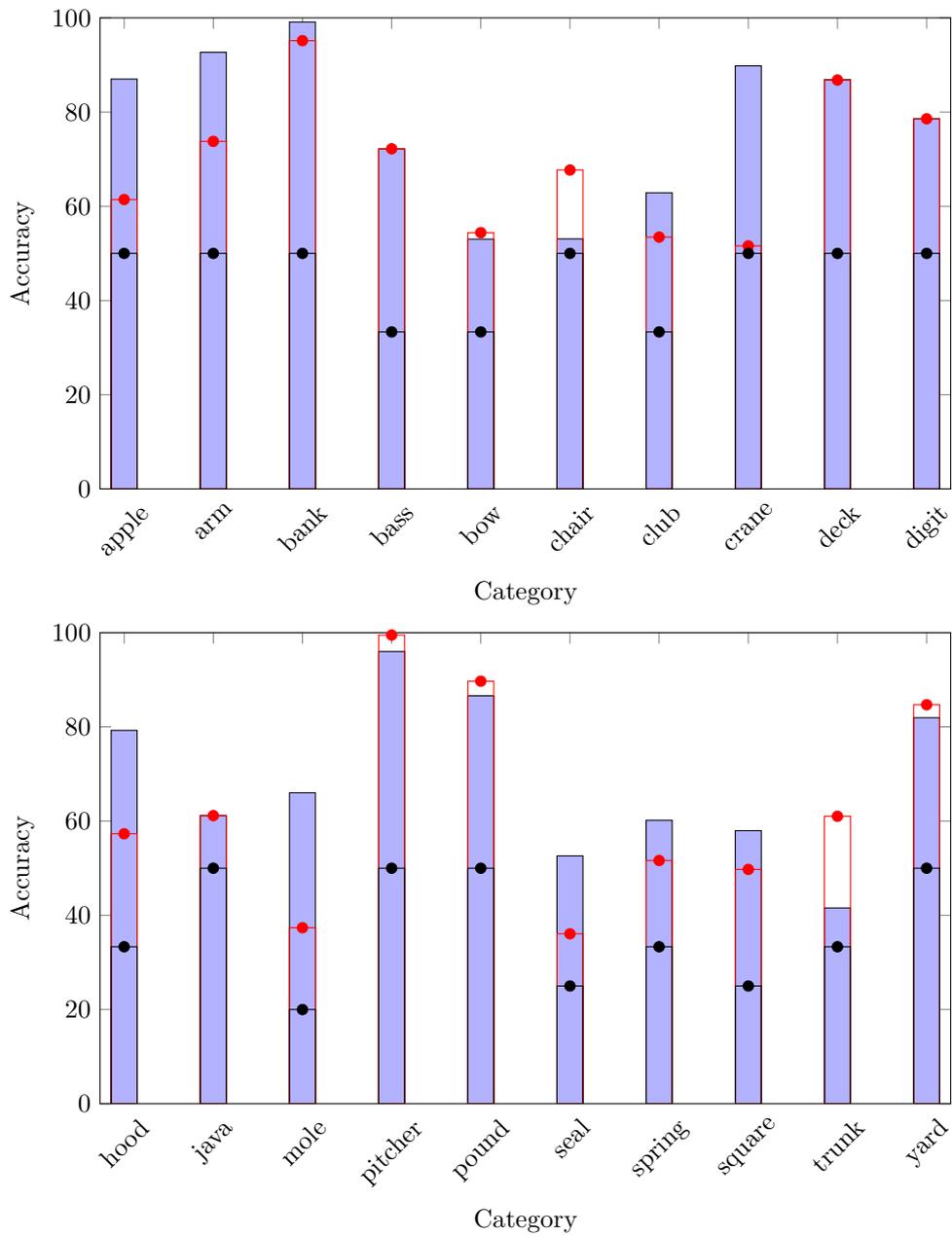

Figure \ref{fig:bert} shows the initial results obtained with the solution implemented by BERT. As can be seen, the results are pretty good since the weak baseline is consistently outperformed, and the strong baseline is almost always outperformed. In addition, there are many use cases where a wide margin beats the strong baseline. Considering that this strategy does not use any external resources or training, this can be considered a good performance.

\begin{figure}
\begin{tikzpicture}
\begin{axis}[bar width=10pt,    ylabel={Accuracy},    xlabel={Category},    xtick=data,    xticklabels={        apple,        arm,        bank,        bass,        bow,        chair,        club,        crane,        deck,        digit    },    ymin=0,    ymax=100,    width=\textwidth,    height=8cm,    legend style={at={(0.5,-0.15)},      anchor=north,legend columns=-1},    enlarge x limits=0.03, xticklabel style={rotate=45 }]

\addplot [ybar, fill=blue!30] coordinates {
    (1,61.43)
    (2,25.70)
    (3,95.16)
    (4,72.25)
    (5,33.95)
    (6,33.80)
    (7,36.67)
    (8,51.56)
    (9,93.00)
    (10,78.57)
};

\addplot[ybar,mark=*,red] coordinates {
    (1,61.43)
    (2,73.78)
    (3,95.16)
    (4,72.20)
    (5,54.4)
    (6,67.7)
    (7,53.46)
    (8,51.59)
    (9,86.8)
    (10,78.57)
};

\addplot[ybar,mark=*,black]  coordinates {
    (1,50)
    (2,50)
    (3,50)
    (4,33.33)
    (5,33.33)
    (6,50)
    (7,33.33)
    (8,50)
    (9,50)
    (10,50)
};
\end{axis}
\end{tikzpicture}

\begin{tikzpicture}
\begin{axis}[ bar width=10pt,    ylabel={Accuracy},    xlabel={Category},    xtick=data,    xticklabels={        hood,        java,        mole,        pitcher,        pound,        seal,        spring,        square,        trunk,        yard    },    ymin=0,    ymax=100,    width=\textwidth,    height=8cm,    legend style={at={(0.5,-0.55)},      anchor=north,legend columns=-1},    enlarge x limits=0.03, , xticklabel style={rotate=45 }   ]
\addplot [ybar, fill=blue!30] coordinates {
    (1,14.63)
    (2,61.17)
    (3,21.85)
    (4,99.53)
    (5,89.70)
    (6,33.33)
    (7,51.64)
    (8,67.63)
    (9,61.03)
    (10,84.72)
};

\addplot[ybar,mark=*,red] coordinates {
    (1,57.31)
    (2,61.17)
    (3,37.37)
    (4,99.5)
    (5,89.7)
    (6,36.08)
    (7,51.64)
    (8,49.75)
    (9,61.02)
    (10,84.7)
};

\addplot[ybar,mark=*,black]  coordinates {
    (1,33.33)
    (2,50)
    (3,20)
    (4,50)
    (5,50)
    (6,25)
    (7,33.33)
    (8,25)
    (9,33.33)
    (10,50)
};
\end{axis}
\end{tikzpicture}
\caption{Results obtained for the CoarseWSD-20 dataset using UWSD+ELMo. The solid blue bar represents the results obtained. While the black and red colors represent the weak and strong baselines, respectively}
\label{fig:elmo}
\end{figure}
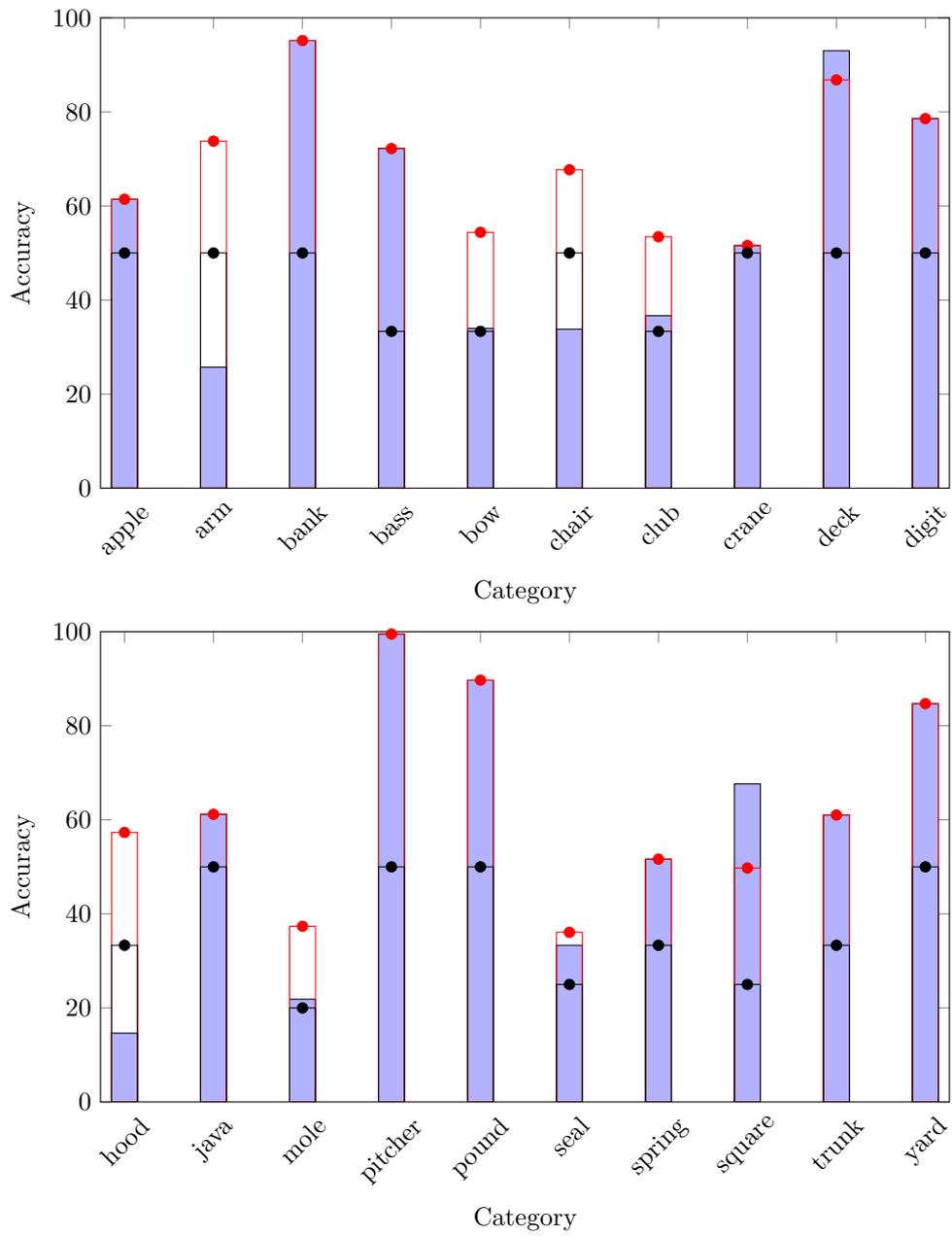

Figure \ref{fig:elmo} shows the results obtained with the solution implemented by ELMo. As can be seen, this approach is better than the weak baseline but often fails to outperform the strong baseline. Therefore, the results are not optimal.

Figure \ref{fig:use} shows the results obtained with the solution implemented by USE embeddings. As can be seen, several results are even lower than those from the weak baseline, and the strong baseline is only surpassed in limited cases. In general, when compared to the other approaches studied, UWSD-USE is not among the best.

\begin{figure}
\begin{tikzpicture}
\begin{axis}[bar width=10pt,    ylabel={Accuracy},    xlabel={Category},    xtick=data,    xticklabels={        apple,        arm,        bank,        bass,        bow,        chair,        club,        crane,        deck,        digit    },    ymin=0,    ymax=100,    width=\textwidth,    height=8cm,    legend style={at={(0.5,-0.15)},      anchor=north,legend columns=-1},    enlarge x limits=0.03, xticklabel style={rotate=45 }]

\addplot [ybar, fill=blue!20] coordinates {
    (1,71.12)
    (2,37.22)
    (3,60.00)
		(4,69.30)
    (5,53.47)
    (6,46.15)
    (7,58.41)
    (8,75.80)
    (9,69.63)
    (10,78.60)
};

\addplot[ybar,mark=*,red] coordinates {
    (1,61.43)
    (2,73.78)
    (3,95.16)
    (4,72.20)
    (5,54.4)
    (6,67.7)
    (7,53.46)
    (8,51.59)
    (9,86.8)
    (10,78.57)
};

\addplot[ybar,mark=*,black]  coordinates {
    (1,50)
    (2,50)
    (3,50)
    (4,33.33)
    (5,33.33)
    (6,50)
    (7,33.33)
    (8,50)
    (9,50)
    (10,50)
};
\end{axis}
\end{tikzpicture}

\begin{tikzpicture}
\begin{axis}[ bar width=10pt,    ylabel={Accuracy},    xlabel={Category},    xtick=data,    xticklabels={        hood,        java,        mole,        pitcher,        pound,        seal,        spring,        square,        trunk,        yard    },    ymin=0,    ymax=100,    width=\textwidth,    height=8cm,    legend style={at={(0.5,-0.55)},      anchor=north,legend columns=-1},    enlarge x limits=0.03, , xticklabel style={rotate=45 }   ]
\addplot [ybar, fill=blue!20] coordinates {
    (1,64.63)
    (2,61.2)
    (3,62.14)
    (4,94.93)
    (5,88.66)
    (6,62.8)
    (7,45.51)
    (8,52.17)
    (9,79.22)
    (10,84.73)
};

\addplot[ybar,mark=*,red] coordinates {
    (1,57.31)
    (2,61.17)
    (3,37.37)
    (4,99.5)
    (5,89.7)
    (6,36.08)
    (7,51.64)
    (8,49.75)
    (9,61.02)
    (10,84.7)
};

\addplot[ybar,mark=*,black]  coordinates {
    (1,33.33)
    (2,50)
    (3,20)
    (4,50)
    (5,50)
    (6,25)
    (7,33.33)
    (8,25)
    (9,33.33)
    (10,50)
};
\end{axis}
\end{tikzpicture}
\caption{Results obtained for the CoarseWSD-20 dataset using UWSD+USE. The solid blue bar represents the results obtained. While the black and red colors represent the weak and strong baselines, respectively}
\label{fig:use}
\end{figure}

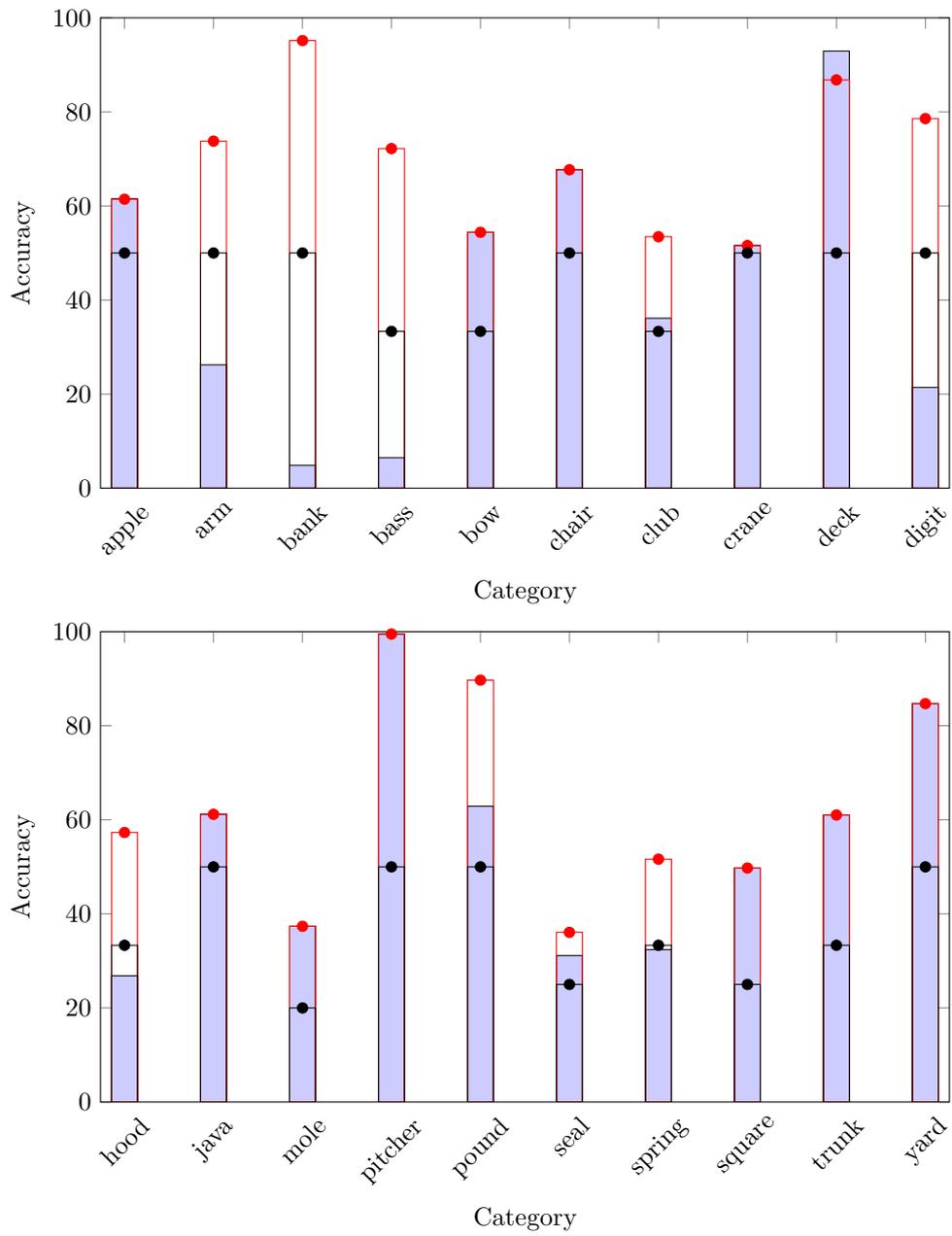
\begin{figure} []
\begin{tikzpicture}
\begin{axis}[bar width=10pt,    ylabel={Accuracy},    xlabel={Category},    xtick=data,    xticklabels={        apple,        arm,        bank,        bass,        bow,        chair,        club,        crane,        deck,        digit    },    ymin=0,    ymax=100,    width=\textwidth,    height=8cm,    legend style={at={(0.5,-0.15)},      anchor=north,legend columns=-1},    enlarge x limits=0.03, xticklabel style={rotate=45 }]

\addplot [ybar, fill=blue!20] coordinates {
    (1,61.53)
    (2,26.22)
    (3,04.85)
    (4,06.47)
    (5,54.42)
    (6,67.70)
    (7,36.13)
    (8,51.60)
    (9,92.93)
    (10,21.43)
};

\addplot[ybar,mark=*,red] coordinates {
    (1,61.43)
    (2,73.78)
    (3,95.16)
    (4,72.20)
    (5,54.4)
    (6,67.7)
    (7,53.46)
    (8,51.59)
    (9,86.8)
    (10,78.57)
};

\addplot[ybar,mark=*,black]  coordinates {
    (1,50)
    (2,50)
    (3,50)
    (4,33.33)
    (5,33.33)
    (6,50)
    (7,33.33)
    (8,50)
    (9,50)
    (10,50)
};
\end{axis}
\end{tikzpicture}

\begin{tikzpicture}
\begin{axis}[ bar width=10pt,    ylabel={Accuracy},    xlabel={Category},    xtick=data,    xticklabels={        hood,        java,        mole,        pitcher,        pound,        seal,        spring,        square,        trunk,        yard    },    ymin=0,    ymax=100,    width=\textwidth,    height=8cm,    legend style={at={(0.5,-0.55)},      anchor=north,legend columns=-1},    enlarge x limits=0.03, , xticklabel style={rotate=45 }   ]
\addplot [ybar, fill=blue!20] coordinates {
    (1,26.83)
    (2,61.2)
    (3,37.4)
    (4,99.53)
    (5,62.89)
    (6,31.13)
    (7,32.39)
    (8,49.76)
    (9,61.04)
    (10,84.73)
};

\addplot[ybar,mark=*,red] coordinates {
    (1,57.31)
    (2,61.17)
    (3,37.37)
    (4,99.5)
    (5,89.7)
    (6,36.08)
    (7,51.64)
    (8,49.75)
    (9,61.02)
    (10,84.7)
};

\addplot[ybar,mark=*,black]  coordinates {
    (1,33.33)
    (2,50)
    (3,20)
    (4,50)
    (5,50)
    (6,25)
    (7,33.33)
    (8,25)
    (9,33.33)
    (10,50)
};
\end{axis}
\end{tikzpicture}
\caption{Results obtained for the CoarseWSD-20 dataset using UWSD+WMD. The solid blue bar represents the results obtained. While the black and red colors represent the weak and strong baselines, respectively}
\label{fig:wmd}
\end{figure}

Figure \ref{fig:wmd} shows the results obtained with the solution implemented by WMD. As can be seen, the results are far from optimal. There are several occasions in which they are even below the weak baseline, being very rare in the cases in which they manage to overcome the strong baseline. Generally speaking, this approach is the one that yields the worst results among those studied.

\subsection{Comparison with existing techniques}
Table \ref{tab:t0} summarizes all our global results. The CoarseWSD-20 dataset is still relatively young and specially designed to perform machine learning, so we are unaware of any other published work on the unsupervised WSD task. However, our experimentation has many alternatives that have been tested. As can be seen, all the proposed strategies can outperform the RO baseline (weak baseline). However, only the strategy that uses BERT embeddings can outperform the MFS baseline (strong baseline) as well. The strategy followed with implementing BERT embeddings is an excellent result since it can outperform a method that uses external knowledge without any extra knowledge or training phase. 

\begin{table}
	\centering
	\scalebox{0.95}{
\begin{tabular}{|c|c|c|}
\hline
Strategy     & Hits & Accuracy \\
\hline
UWSD+BERT  & 7,927    &   77.74\%   \\
MFS-Baseline & 7,487    &   73.43\%    \\
UWSD+USE  & 7,335    &   71.94\%    \\
UWSD+ELMo & 7,010    &   68.75\%   \\
UWSD+WMD  & 6,123    &   60.00\%     \\
RO-Baseline & 4,459    &   43.73\%    \\
\hline     
\end{tabular}}
	\caption{Summary of the best results obtained for each of the different embedding approaches}
	\label{tab:t0}
\end{table}

\subsection{Detailed Results}
In the following, we show the results of all the experiments carried out, i.e., taking into account all the language models that have been analyzed and the results they have led to

Table \ref{tab:t1} summarizes all the results obtained with different BERT models. All these models have been extensively evaluated for their quality of embedded sentences. As can be seen, not all models lead to results superior to the baseline. A more in-depth analysis of why some models considered can rank better than others remains a future work in progress.

\begin{table}[h!]
	\centering
	\scalebox{0.95}{
\begin{tabular}{|c|c|c|}
\hline
Strategy     & Hits & Accuracy \\
\hline
UWSD+BERT+all-mpnet-base-v2  & 7,927    &   77.74\%   \\
UWSD+BERT+all-MiniLM-L12-v2 & 7,652    &   75.05\%   \\
UWSD+BERT+all-MiniLM-L6-v2  & 7,609    &   74.63\%   \\
MFS-Baseline & 7,487    &   73.43\%    \\
UWSD+BERT+paraphrase-albert-small-v2 & 7,104  & 69.67\%   \\
UWSD+BERT+paraphrase-MiniLM-L3-v2 & 7,098    &   69.62\%    \\
UWSD+BERT+all-distilroberta-v1  & 5,547    &   54.40\%   \\
RO-Baseline & 4,459    &   43.73\%    \\
\hline     
\end{tabular}}
	\caption{Summary of the results results obtained using different language models based on BERT}
	\label{tab:t1}
\end{table}

Table \ref{tab:t2} summarizes all the results obtained with different ELMo models. These models have undergone, once again, thorough assessments to determine the quality of the ELMo embeddings. No model surpasses the baseline in terms of results.

\begin{table}[h!]
	\centering
	\scalebox{0.95}{
\begin{tabular}{|c|c|c|}
\hline
Strategy     & Hits & Accuracy \\
\hline
MFS-Baseline & 7,487    &   73.43\%    \\
UWSD+ELMo+Corpus of Historical American English & 7,010    &   68.75\%   \\ 
UWSD+ELMo+English Wikipedia February 2017 & 5,593    &   54.85\%   \\
UWSD+ELMo+English Wikipedia October 2019 & 4,786   &   46.94\%   \\
RO-Baseline & 4,459    &   43.73\%    \\
\hline     
\end{tabular}}
	\caption{Summary of the results obtained using different language models based on ELMo}
	\label{tab:t2}
\end{table}

Table \ref{tab:t3} summarizes all the results of different USE models. These models have been evaluated once more to assess the quality of the USE embeddings. The evaluations indicate that none of the models exceed the baseline in terms of performance, although the Large model achieves results very close to the baseline.

\begin{table}[h!]
	\centering
	\scalebox{0.95}{
\begin{tabular}{|c|c|c|}
\hline
Strategy     & Hits & Accuracy \\
\hline
MFS-Baseline & 7,487    &   73.43\%    \\
UWSD+USE+Large & 7,335    &   71.94\%   \\
UWSD+USE+Classic & 6,396    &   62.73\%    \\ 
RO-Baseline & 4,459    &   43.73\%    \\
\hline     
\end{tabular}}
	\caption{Summary of the results results obtained using different language models based on USE}
	\label{tab:t3}
\end{table}

Lastly, Table \ref{tab:t4} summarizes all the results obtained with different WSD models. The models were again evaluated to gauge the embedding's quality. The findings show that none of the models outperform the established baseline regarding outcomes.

\begin{table}[h!]
	\centering
	\scalebox{0.95}{
\begin{tabular}{|c|c|c|}
\hline
Strategy     & Hits & Accuracy \\
\hline
MFS-Baseline & 7,487    &   73.43\%    \\
UWSD+WMD+glove-twitter-200 & 6,123    &   60.00\%   \\
UWSD+WMD+word2vec-google-news-300 & 5,868    &   57.55\%  \\ 
UWSD+WMD+glove-wiki-gigaword-300 & 5,858    &   57.45\%   \\
UWSD+WMD+fasttext-wiki-news-subwords-300 & 5,847    &   57.34\%    \\
RO-Baseline & 4,459    &   43.73\%    \\
\hline     
\end{tabular}}
	\caption{Summary of the results results obtained using different language models based on WMD}
	\label{tab:t4}
\end{table}

\section{Discussion}
Our strategy has exhibited promising results in improving the accuracy of WSD since it can effectively capture the subtle distinctions in word senses, leading to more precise disambiguation. One advantage of the proposed strategy is its ability to function without annotated data, making it more widely applicable to diverse domains. This is particularly crucial for low-resource languages where annotated data is scarce.

Another advantage of our method is its capability to capture the complex nuances of meaning that traditional semantic models might overlook. The rationale behind incorporating contextual information is to differentiate between polysemous words with various meanings in different contexts. 

The experimental results demonstrate that the proposed method outperforms several unsupervised WSD techniques on the CoarseWSD-20 benchmark dataset. The performance improvement is particularly remarkable for words with high levels of ambiguity, where conventional methods often struggle to disambiguate accurately.

While the proposed approach displays promise, some limitations still require addressing. One limitation is that the approach heavily relies on the quality of available contextual information. Disambiguation accuracy may be compromised when the context is noisy or ambiguous. Additionally, the proposed approach may perform inadequately in cases where the context is too sparse, resulting in inadequate information for precise disambiguation. Future research directions could explore further improvements to the approach, such as integrating additional sources of contextual information or combining it with other WSD techniques.

\section{Conclusion}
This work shows how CASS can comprehend human language effectively since it can recognize that the meaning of a word can differ based on the specific context in which it appears. CASS techniques aim to capture this variability, calculate more accurate similarity scores, and improve the performance of unsupervised WSD strategies.

We have seen that including contextual information is a proven way to improve accuracy when facing unsupervised WSD tasks. Our research indicates that using a strategy of this kind can address the challenge of interpreting the meaning of language as used in particular settings. We have achieved significant improvements in disambiguation accuracy compared to conventional methods that do not consider contextual information. The fact that our strategy performs better than numerous other unsupervised strategies is further evidence of its usefulness in dealing with the ambiguity of word senses.

In conclusion, CASS and WSD have a great deal of untapped potential that may improve the accuracy of text understanding and inspire the development of novel NLP applications that use disambiguation technology. Novel strategies in this direction could lead to developing more efficient solutions and better comprehending the complexities and nuances of human language.

\section*{Acknowledgments}
This research has been funded by the Federal Ministry for Climate Action, Environment, Energy, Mobility, Innovation, and Technology (BMK), the Federal Ministry for Digital and Economic Affairs (BMDW), and the State of Upper Austria in the frame of SCCH, a center in the COMET - Competence Centers for Excellent Technologies Programme managed by Austrian Research Promotion Agency FFG.

\bibliography{mybib}


\begin{thebibliography}{30}
\ifx \bisbn   \undefined \def \bisbn  #1{ISBN #1}\fi
\ifx \binits  \undefined \def \binits#1{#1}\fi
\ifx \bauthor  \undefined \def \bauthor#1{#1}\fi
\ifx \batitle  \undefined \def \batitle#1{#1}\fi
\ifx \bjtitle  \undefined \def \bjtitle#1{#1}\fi
\ifx \bvolume  \undefined \def \bvolume#1{\textbf{#1}}\fi
\ifx \byear  \undefined \def \byear#1{#1}\fi
\ifx \bissue  \undefined \def \bissue#1{#1}\fi
\ifx \bfpage  \undefined \def \bfpage#1{#1}\fi
\ifx \blpage  \undefined \def \blpage #1{#1}\fi
\ifx \burl  \undefined \def \burl#1{\textsf{#1}}\fi
\ifx \doiurl  \undefined \def \doiurl#1{\url{https://doi.org/#1}}\fi
\ifx \betal  \undefined \def \betal{\textit{et al.}}\fi
\ifx \binstitute  \undefined \def \binstitute#1{#1}\fi
\ifx \binstitutionaled  \undefined \def \binstitutionaled#1{#1}\fi
\ifx \bctitle  \undefined \def \bctitle#1{#1}\fi
\ifx \beditor  \undefined \def \beditor#1{#1}\fi
\ifx \bpublisher  \undefined \def \bpublisher#1{#1}\fi
\ifx \bbtitle  \undefined \def \bbtitle#1{#1}\fi
\ifx \bedition  \undefined \def \bedition#1{#1}\fi
\ifx \bseriesno  \undefined \def \bseriesno#1{#1}\fi
\ifx \blocation  \undefined \def \blocation#1{#1}\fi
\ifx \bsertitle  \undefined \def \bsertitle#1{#1}\fi
\ifx \bsnm \undefined \def \bsnm#1{#1}\fi
\ifx \bsuffix \undefined \def \bsuffix#1{#1}\fi
\ifx \bparticle \undefined \def \bparticle#1{#1}\fi
\ifx \barticle \undefined \def \barticle#1{#1}\fi
\bibcommenthead
\ifx \bconfdate \undefined \def \bconfdate #1{#1}\fi
\ifx \botherref \undefined \def \botherref #1{#1}\fi
\ifx \url \undefined \def \url#1{\textsf{#1}}\fi
\ifx \bchapter \undefined \def \bchapter#1{#1}\fi
\ifx \bbook \undefined \def \bbook#1{#1}\fi
\ifx \bcomment \undefined \def \bcomment#1{#1}\fi
\ifx \oauthor \undefined \def \oauthor#1{#1}\fi
\ifx \citeauthoryear \undefined \def \citeauthoryear#1{#1}\fi
\ifx \endbibitem  \undefined \def \endbibitem {}\fi
\ifx \bconflocation  \undefined \def \bconflocation#1{#1}\fi
\ifx \arxivurl  \undefined \def \arxivurl#1{\textsf{#1}}\fi
\csname PreBibitemsHook\endcsname

\bibitem[\protect\citeauthoryear{Navigli and Martelli}{2019}]{key-Navigli}
\begin{barticle}
\bauthor{\bsnm{Navigli}, \binits{R.}},
\bauthor{\bsnm{Martelli}, \binits{F.}}:
\batitle{An overview of word and sense similarity}.
\bjtitle{Nat. Lang. Eng.}
\bvolume{25}(\bissue{6}),
\bfpage{693}--\blpage{714}
(\byear{2019})
\doiurl{10.1017/S1351324919000305}
\end{barticle}
\endbibitem

\bibitem[\protect\citeauthoryear{Lastra{-}D{\'{\i}}az
  et~al.}{2017}]{key-Lastra-HESML}
\begin{barticle}
\bauthor{\bsnm{Lastra{-}D{\'{\i}}az}, \binits{J.J.}},
\bauthor{\bsnm{Garc{\'{\i}}a{-}Serrano}, \binits{A.}},
\bauthor{\bsnm{Batet}, \binits{M.}},
\bauthor{\bsnm{Fern{\'{a}}ndez}, \binits{M.}},
\bauthor{\bsnm{Chirigati}, \binits{F.}}:
\batitle{{HESML:} {A} scalable ontology-based semantic similarity measures
  library with a set of reproducible experiments and a replication dataset}.
\bjtitle{Inf. Syst.}
\bvolume{66},
\bfpage{97}--\blpage{118}
(\byear{2017})
\doiurl{10.1016/j.is.2017.02.002}
\end{barticle}
\endbibitem

\bibitem[\protect\citeauthoryear{Martinez-Gil}{2022}]{key-martinez-mlwa}
\begin{barticle}
\bauthor{\bsnm{Martinez-Gil}, \binits{J.}}:
\batitle{A comprehensive review of stacking methods for semantic similarity
  measurement}.
\bjtitle{Machine Learning with Applications}
\bvolume{10},
\bfpage{100423}
(\byear{2022})
\doiurl{10.1016/j.mlwa.2022.100423}
\end{barticle}
\endbibitem

\bibitem[\protect\citeauthoryear{Navigli}{2009}]{navigli2009word}
\begin{barticle}
\bauthor{\bsnm{Navigli}, \binits{R.}}:
\batitle{Word sense disambiguation: A survey}.
\bjtitle{ACM computing surveys (CSUR)}
\bvolume{41}(\bissue{2}),
\bfpage{1}--\blpage{69}
(\byear{2009})
\doiurl{10.1145/1459352.1459355}
\end{barticle}
\endbibitem

\bibitem[\protect\citeauthoryear{Han and Shirai}{2021}]{key-Han}
\begin{bchapter}
\bauthor{\bsnm{Han}, \binits{S.}},
\bauthor{\bsnm{Shirai}, \binits{K.}}:
\bctitle{Unsupervised word sense disambiguation based on word embedding and
  collocation}.
In: \beditor{\bsnm{Rocha}, \binits{A.P.}},
\beditor{\bsnm{Steels}, \binits{L.}},
\beditor{\bsnm{Herik}, \binits{H.J.}} (eds.)
\bbtitle{Proceedings of the 13th International Conference on Agents and
  Artificial Intelligence, {ICAART} 2021, Volume 2, Online Streaming, February
  4-6, 2021},
pp. \bfpage{1218}--\blpage{1225}.
\bpublisher{SCITEPRESS},
\blocation{Online Streaming}
(\byear{2021}).
\doiurl{10.5220/0010380112181225}
\end{bchapter}
\endbibitem

\bibitem[\protect\citeauthoryear{Moradi et~al.}{2019}]{key-Moradi}
\begin{bchapter}
\bauthor{\bsnm{Moradi}, \binits{B.}},
\bauthor{\bsnm{Ansari}, \binits{E.}},
\bauthor{\bsnm{Zabokrtsk{\'{y}}}, \binits{Z.}}:
\bctitle{Unsupervised word sense disambiguation using word embeddings}.
In: \bbtitle{25th Conference of Open Innovations Association, {FRUCT} 2019,
  November 5-8, 2019},
pp. \bfpage{228}--\blpage{233}.
\bpublisher{{IEEE}},
\blocation{Helsinki, Finland}
(\byear{2019}).
\doiurl{10.23919/FRUCT48121.2019.8981526}
\end{bchapter}
\endbibitem

\bibitem[\protect\citeauthoryear{Rahman and Borah}{2022}]{key-Rahman}
\begin{barticle}
\bauthor{\bsnm{Rahman}, \binits{N.}},
\bauthor{\bsnm{Borah}, \binits{B.}}:
\batitle{An unsupervised method for word sense disambiguation}.
\bjtitle{J. King Saud Univ. Comput. Inf. Sci.}
\bvolume{34}(\bissue{9}),
\bfpage{6643}--\blpage{6651}
(\byear{2022})
\doiurl{10.1016/j.jksuci.2021.07.022}
\end{barticle}
\endbibitem

\bibitem[\protect\citeauthoryear{Ustalov et~al.}{2018}]{key-Ustalov}
\begin{bchapter}
\bauthor{\bsnm{Ustalov}, \binits{D.}},
\bauthor{\bsnm{Teslenko}, \binits{D.}},
\bauthor{\bsnm{Panchenko}, \binits{A.}},
\bauthor{\bsnm{Chernoskutov}, \binits{M.}},
\bauthor{\bsnm{Biemann}, \binits{C.}},
\bauthor{\bsnm{Ponzetto}, \binits{S.P.}}:
\bctitle{An unsupervised word sense disambiguation system for under-resourced
  languages}.
In: \bbtitle{Proceedings of the Eleventh International Conference on Language
  Resources and Evaluation, {LREC} 2018, May 7-12, 2018}.
\bpublisher{European Language Resources Association {(ELRA)}},
\blocation{Miyazaki, Japan}
(\byear{2018})
\end{bchapter}
\endbibitem

\bibitem[\protect\citeauthoryear{Han et~al.}{2013}]{key-sts}
\begin{bchapter}
\bauthor{\bsnm{Han}, \binits{L.}},
\bauthor{\bsnm{Kashyap}, \binits{A.L.}},
\bauthor{\bsnm{Finin}, \binits{T.}},
\bauthor{\bsnm{Mayfield}, \binits{J.}},
\bauthor{\bsnm{Weese}, \binits{J.}}:
\bctitle{Umbc{\_}ebiquity-core: Semantic textual similarity systems}.
In: \beditor{\bsnm{Diab}, \binits{M.T.}},
\beditor{\bsnm{Baldwin}, \binits{T.}},
\beditor{\bsnm{Baroni}, \binits{M.}} (eds.)
\bbtitle{Proceedings of the Second Joint Conference on Lexical and
  Computational Semantics, *SEM 2013, June 13-14, 2013},
pp. \bfpage{44}--\blpage{52}.
\bpublisher{Association for Computational Linguistics},
\blocation{Atlanta, Georgia (USA)}
(\byear{2013})
\end{bchapter}
\endbibitem

\bibitem[\protect\citeauthoryear{Harispe et~al.}{2015}]{key-Harispe}
\begin{bbook}
\bauthor{\bsnm{Harispe}, \binits{S.}},
\bauthor{\bsnm{Ranwez}, \binits{S.}},
\bauthor{\bsnm{Janaqi}, \binits{S.}},
\bauthor{\bsnm{Montmain}, \binits{J.}}:
\bbtitle{Semantic Similarity from Natural Language and Ontology Analysis}.
\bsertitle{Synthesis Lectures on Human Language Technologies}.
\bpublisher{Morgan {\&} Claypool Publishers},
\blocation{-}
(\byear{2015}).
\doiurl{10.2200/S00639ED1V01Y201504HLT027}
\end{bbook}
\endbibitem

\bibitem[\protect\citeauthoryear{Lastra{-}D{\'{\i}}az and
  Garc{\'{\i}}a{-}Serrano}{2015}]{key-Lastra3}
\begin{barticle}
\bauthor{\bsnm{Lastra{-}D{\'{\i}}az}, \binits{J.J.}},
\bauthor{\bsnm{Garc{\'{\i}}a{-}Serrano}, \binits{A.}}:
\batitle{A new family of information content models with an experimental survey
  on wordnet}.
\bjtitle{Knowl.-Based Syst.}
\bvolume{89},
\bfpage{509}--\blpage{526}
(\byear{2015})
\doiurl{10.1016/j.knosys.2015.08.019}
\end{barticle}
\endbibitem

\bibitem[\protect\citeauthoryear{Lastra{-}D{\'{\i}}az
  et~al.}{2019}]{key-Lastra}
\begin{barticle}
\bauthor{\bsnm{Lastra{-}D{\'{\i}}az}, \binits{J.J.}},
\bauthor{\bsnm{Goikoetxea}, \binits{J.}},
\bauthor{\bsnm{Taieb}, \binits{M.A.H.}},
\bauthor{\bsnm{Garc{\'{\i}}a{-}Serrano}, \binits{A.}},
\bauthor{\bsnm{Aouicha}, \binits{M.B.}},
\bauthor{\bsnm{Agirre}, \binits{E.}}:
\batitle{A reproducible survey on word embeddings and ontology-based methods
  for word similarity: Linear combinations outperform the state of the art}.
\bjtitle{Eng. Appl. Artif. Intell.}
\bvolume{85},
\bfpage{645}--\blpage{665}
(\byear{2019})
\doiurl{10.1016/j.engappai.2019.07.010}
\end{barticle}
\endbibitem

\bibitem[\protect\citeauthoryear{Martinez-Gil and
  Chaves{-}Gonzalez}{2021}]{key-martinez-kbs}
\begin{barticle}
\bauthor{\bsnm{Martinez-Gil}, \binits{J.}},
\bauthor{\bsnm{Chaves{-}Gonzalez}, \binits{J.M.}}:
\batitle{Semantic similarity controllers: On the trade-off between accuracy and
  interpretability}.
\bjtitle{Knowl. Based Syst.}
\bvolume{234},
\bfpage{107609}
(\byear{2021})
\doiurl{10.1016/j.knosys.2021.107609}
\end{barticle}
\endbibitem

\bibitem[\protect\citeauthoryear{Zhu and Iglesias}{2017}]{key-sematch}
\begin{barticle}
\bauthor{\bsnm{Zhu}, \binits{G.}},
\bauthor{\bsnm{Iglesias}, \binits{C.A.}}:
\batitle{Computing semantic similarity of concepts in knowledge graphs}.
\bjtitle{{IEEE} Trans. Knowl. Data Eng.}
\bvolume{29}(\bissue{1}),
\bfpage{72}--\blpage{85}
(\byear{2017})
\doiurl{10.1109/TKDE.2016.2610428}
\end{barticle}
\endbibitem

\bibitem[\protect\citeauthoryear{Martinez-Gil and
  Chaves-Gonzalez}{2022}]{key-martinez-jfis}
\begin{barticle}
\bauthor{\bsnm{Martinez-Gil}, \binits{J.}},
\bauthor{\bsnm{Chaves-Gonzalez}, \binits{J.M.}}:
\batitle{Sustainable semantic similarity assessment}.
\bjtitle{Journal of Intelligent \& Fuzzy Systems}
\bvolume{43}(\bissue{5}),
\bfpage{6163}--\blpage{6174}
(\byear{2022})
\doiurl{10.3233/JIFS-220137}
\end{barticle}
\endbibitem

\bibitem[\protect\citeauthoryear{Martinez-Gil and
  Chaves-Gonzalez}{2020}]{key-martinez-eswa2}
\begin{barticle}
\bauthor{\bsnm{Martinez-Gil}, \binits{J.}},
\bauthor{\bsnm{Chaves-Gonzalez}, \binits{J.M.}}:
\batitle{A novel method based on symbolic regression for interpretable semantic
  similarity measurement}.
\bjtitle{Expert Syst. Appl.}
\bvolume{160},
\bfpage{113663}
(\byear{2020})
\doiurl{10.1016/j.eswa.2020.113663}
\end{barticle}
\endbibitem

\bibitem[\protect\citeauthoryear{Bollegala et~al.}{2011}]{key-Bollegala}
\begin{barticle}
\bauthor{\bsnm{Bollegala}, \binits{D.}},
\bauthor{\bsnm{Matsuo}, \binits{Y.}},
\bauthor{\bsnm{Ishizuka}, \binits{M.}}:
\batitle{A web search engine-based approach to measure semantic similarity
  between words}.
\bjtitle{{IEEE} Trans. Knowl. Data Eng.}
\bvolume{23}(\bissue{7}),
\bfpage{977}--\blpage{990}
(\byear{2011})
\doiurl{10.1109/TKDE.2010.172}
\end{barticle}
\endbibitem

\bibitem[\protect\citeauthoryear{Pilehvar and Navigli}{2015}]{key-Pilehvar}
\begin{barticle}
\bauthor{\bsnm{Pilehvar}, \binits{M.T.}},
\bauthor{\bsnm{Navigli}, \binits{R.}}:
\batitle{From senses to texts: An all-in-one graph-based approach for measuring
  semantic similarity}.
\bjtitle{Artif. Intell.}
\bvolume{228},
\bfpage{95}--\blpage{128}
(\byear{2015})
\doiurl{10.1016/j.artint.2015.07.005}
\end{barticle}
\endbibitem

\bibitem[\protect\citeauthoryear{Chandrasekaran and
  Mago}{2021}]{key-chandrasekaran}
\begin{barticle}
\bauthor{\bsnm{Chandrasekaran}, \binits{D.}},
\bauthor{\bsnm{Mago}, \binits{V.}}:
\batitle{Evolution of semantic similarity - {A} survey}.
\bjtitle{{ACM} Comput. Surv.}
\bvolume{54}(\bissue{2}),
\bfpage{41}--\blpage{14137}
(\byear{2021})
\doiurl{10.1145/3440755}
\end{barticle}
\endbibitem

\bibitem[\protect\citeauthoryear{Apidianaki}{2022}]{key-apidianaki}
\begin{botherref}
\oauthor{\bsnm{Apidianaki}, \binits{M.}}:
From word types to tokens and back: A survey of approaches to word meaning
  representation and interpretation.
Computational Linguistics,
1--60
(2022)
\doiurl{10.1162/coli_a_00474}
\end{botherref}
\endbibitem

\bibitem[\protect\citeauthoryear{Loureiro et~al.}{2022}]{key-loureiro2022lmms}
\begin{barticle}
\bauthor{\bsnm{Loureiro}, \binits{D.}},
\bauthor{\bsnm{Jorge}, \binits{A.M.}},
\bauthor{\bsnm{Camacho-Collados}, \binits{J.}}:
\batitle{Lmms reloaded: Transformer-based sense embeddings for disambiguation
  and beyond}.
\bjtitle{Artificial Intelligence}
\bvolume{305},
\bfpage{103661}
(\byear{2022})
\end{barticle}
\endbibitem

\bibitem[\protect\citeauthoryear{Eyal et~al.}{2022}]{key-eyal}
\begin{botherref}
\oauthor{\bsnm{Eyal}, \binits{M.}},
\oauthor{\bsnm{Sadde}, \binits{S.}},
\oauthor{\bsnm{Taub{-}Tabib}, \binits{H.}},
\oauthor{\bsnm{Goldberg}, \binits{Y.}}:
Large scale substitution-based word sense induction,
4738--4752
(2022)
\doiurl{10.18653/V1/2022.ACL-LONG.325}
\end{botherref}
\endbibitem

\bibitem[\protect\citeauthoryear{Devlin et~al.}{2019}]{key-Bert}
\begin{bchapter}
\bauthor{\bsnm{Devlin}, \binits{J.}},
\bauthor{\bsnm{Chang}, \binits{M.}},
\bauthor{\bsnm{Lee}, \binits{K.}},
\bauthor{\bsnm{Toutanova}, \binits{K.}}:
\bctitle{{BERT:} pre-training of deep bidirectional transformers for language
  understanding}.
In: \beditor{\bsnm{Burstein}, \binits{J.}},
\beditor{\bsnm{Doran}, \binits{C.}},
\beditor{\bsnm{Solorio}, \binits{T.}} (eds.)
\bbtitle{Proceedings of the 2019 Conference of the North American Chapter of
  the Association for Computational Linguistics: Human Language Technologies,
  {NAACL-HLT} 2019, June 2-7, 2019, Volume 1 (Long and Short Papers)},
pp. \bfpage{4171}--\blpage{4186}.
\bpublisher{Association for Computational Linguistics},
\blocation{Minneapolis, MN, USA}
(\byear{2019}).
\doiurl{10.18653/v1/n19-1423}
\end{bchapter}
\endbibitem

\bibitem[\protect\citeauthoryear{Peters et~al.}{2018}]{key-elmo}
\begin{bchapter}
\bauthor{\bsnm{Peters}, \binits{M.E.}},
\bauthor{\bsnm{Neumann}, \binits{M.}},
\bauthor{\bsnm{Iyyer}, \binits{M.}},
\bauthor{\bsnm{Gardner}, \binits{M.}},
\bauthor{\bsnm{Clark}, \binits{C.}},
\bauthor{\bsnm{Lee}, \binits{K.}},
\bauthor{\bsnm{Zettlemoyer}, \binits{L.}}:
\bctitle{Deep contextualized word representations}.
In: \beditor{\bsnm{Walker}, \binits{M.A.}},
\beditor{\bsnm{Ji}, \binits{H.}},
\beditor{\bsnm{Stent}, \binits{A.}} (eds.)
\bbtitle{Proceedings of the 2018 Conference of the North American Chapter of
  the Association for Computational Linguistics: Human Language Technologies,
  {NAACL-HLT} 2018, June 1-6, 2018, Volume 1 (Long Papers)},
pp. \bfpage{2227}--\blpage{2237}.
\bpublisher{Association for Computational Linguistics},
\blocation{New Orleans, Louisiana, USA}
(\byear{2018}).
\doiurl{10.18653/v1/n18-1202}
\end{bchapter}
\endbibitem

\bibitem[\protect\citeauthoryear{Cer et~al.}{2018}]{key-cer}
\begin{bchapter}
\bauthor{\bsnm{Cer}, \binits{D.}},
\bauthor{\bsnm{Yang}, \binits{Y.}},
\bauthor{\bsnm{Kong}, \binits{S.}},
\bauthor{\bsnm{Hua}, \binits{N.}},
\bauthor{\bsnm{Limtiaco}, \binits{N.}},
\bauthor{\bsnm{John}, \binits{R.S.}},
\bauthor{\bsnm{Constant}, \binits{N.}},
\bauthor{\bsnm{Guajardo{-}Cespedes}, \binits{M.}},
\bauthor{\bsnm{Yuan}, \binits{S.}},
\bauthor{\bsnm{Tar}, \binits{C.}},
\bauthor{\bsnm{Strope}, \binits{B.}},
\bauthor{\bsnm{Kurzweil}, \binits{R.}}:
\bctitle{Universal sentence encoder for english}.
In: \beditor{\bsnm{Blanco}, \binits{E.}},
\beditor{\bsnm{Lu}, \binits{W.}} (eds.)
\bbtitle{Proceedings of the 2018 Conference on Empirical Methods in Natural
  Language Processing, {EMNLP} 2018: System Demonstrations, October 31 -
  November 4, 2018},
pp. \bfpage{169}--\blpage{174}.
\bpublisher{Association for Computational Linguistics},
\blocation{Brussels, Belgium}
(\byear{2018}).
\doiurl{10.18653/v1/d18-2029}
\end{bchapter}
\endbibitem

\bibitem[\protect\citeauthoryear{Kusner et~al.}{2015}]{key-kusner}
\begin{bchapter}
\bauthor{\bsnm{Kusner}, \binits{M.}},
\bauthor{\bsnm{Sun}, \binits{Y.}},
\bauthor{\bsnm{Kolkin}, \binits{N.}},
\bauthor{\bsnm{Weinberger}, \binits{K.}}:
\bctitle{From word embeddings to document distances}.
In: \bbtitle{International Conference on Machine Learning},
pp. \bfpage{957}--\blpage{966}
(\byear{2015}).
\bcomment{PMLR}
\end{bchapter}
\endbibitem

\bibitem[\protect\citeauthoryear{Martinez-Gil
  et~al.}{2021}]{key-martinez-dawak}
\begin{bchapter}
\bauthor{\bsnm{Martinez-Gil}, \binits{J.}},
\bauthor{\bsnm{Mokadem}, \binits{R.}},
\bauthor{\bsnm{K{\"{u}}ng}, \binits{J.}},
\bauthor{\bsnm{Hameurlain}, \binits{A.}}:
\bctitle{A novel neurofuzzy approach for semantic similarity measurement}.
In: \beditor{\bsnm{Golfarelli}, \binits{M.}},
\beditor{\bsnm{Wrembel}, \binits{R.}},
\beditor{\bsnm{Kotsis}, \binits{G.}},
\beditor{\bsnm{Tjoa}, \binits{A.M.}},
\beditor{\bsnm{Khalil}, \binits{I.}} (eds.)
\bbtitle{Big Data Analytics and Knowledge Discovery - 23rd International
  Conference, DaWaK 2021, Virtual Event, September 27-30, 2021, Proceedings}.
\bsertitle{Lecture Notes in Computer Science},
vol. \bseriesno{12925},
pp. \bfpage{192}--\blpage{203}.
\bpublisher{Springer},
\blocation{Virtual Event}
(\byear{2021}).
\doiurl{10.1007/978-3-030-86534-4\_18}
\end{bchapter}
\endbibitem

\bibitem[\protect\citeauthoryear{Mikolov et~al.}{2013}]{key-Mikolov}
\begin{bchapter}
\bauthor{\bsnm{Mikolov}, \binits{T.}},
\bauthor{\bsnm{Sutskever}, \binits{I.}},
\bauthor{\bsnm{Chen}, \binits{K.}},
\bauthor{\bsnm{Corrado}, \binits{G.S.}},
\bauthor{\bsnm{Dean}, \binits{J.}}:
\bctitle{Distributed representations of words and phrases and their
  compositionality}.
In: \bbtitle{Advances in Neural Information Processing Systems 26: 27th Annual
  Conference on Neural Information Processing Systems 2013. Proceedings of a
  Meeting Held December 5-8, 2013, Lake Tahoe, Nevada, United States.},
pp. \bfpage{3111}--\blpage{3119}
(\byear{2013})
\end{bchapter}
\endbibitem

\bibitem[\protect\citeauthoryear{Skianis et~al.}{2020}]{skianis2020boosting}
\begin{bchapter}
\bauthor{\bsnm{Skianis}, \binits{K.}},
\bauthor{\bsnm{Malliaros}, \binits{F.D.}},
\bauthor{\bsnm{Tziortziotis}, \binits{N.}},
\bauthor{\bsnm{Vazirgiannis}, \binits{M.}}:
\bctitle{Boosting tricks for word mover’s distance}.
In: \bbtitle{International Conference on Artificial Neural Networks},
pp. \bfpage{761}--\blpage{772}
(\byear{2020}).
\doiurl{10.1007/978-3-030-61616-8\_61} .
\bcomment{Springer}
\end{bchapter}
\endbibitem

\bibitem[\protect\citeauthoryear{Loureiro et~al.}{2021}]{key-loureiro}
\begin{barticle}
\bauthor{\bsnm{Loureiro}, \binits{D.}},
\bauthor{\bsnm{Rezaee}, \binits{K.}},
\bauthor{\bsnm{Pilehvar}, \binits{M.T.}},
\bauthor{\bsnm{Camacho-Collados}, \binits{J.}}:
\batitle{Analysis and evaluation of language models for word sense
  disambiguation}.
\bjtitle{Computational Linguistics}
\bvolume{47}(\bissue{2}),
\bfpage{387}--\blpage{443}
(\byear{2021})
\doiurl{10.1162/coli_a_00405}
\end{barticle}
\endbibitem

\end{thebibliography}

\end{document}